\documentclass[conference]{IEEEtran}
\IEEEoverridecommandlockouts
\usepackage{cite}
\usepackage{amsmath,amssymb,amsfonts}
\usepackage{algorithmic}
\usepackage{graphicx}
\usepackage{textcomp}
\usepackage{xcolor}
\usepackage{array}
\usepackage{verbatim}
\usepackage{booktabs}
\usepackage{multirow}
\usepackage{makecell}
\usepackage{caption}
\usepackage[table]{xcolor}
\usepackage{textcomp}
\usepackage{stfloats}

\def\BibTeX{{\rm B\kern-.05em{\sc i\kern-.025em b}\kern-.08em
    T\kern-.1667em\lower.7ex\hbox{E}\kern-.125emX}}
\begin{document}

\title{SeVeDo: A Heterogeneous Transformer Accelerator \\ for Low-Bit Inference via Hierarchical Group Quantization and SVD-Guided Mixed Precision}

\author{\IEEEauthorblockN{Yuseon Choi, Sangjin Kim, Jungjun Oh, Byeongcheol Kim, and Hoi-Jun Yoo}
\IEEEauthorblockA{\textit{Korea Advanced Institute of Science and Technology (KAIST)} \\
Daejeon, Republic of Korea \\
yuseon.choi@kaist.ac.kr}
}

\maketitle

\begin{abstract}
Low-bit quantization is a promising technique for efficient transformer inference by reducing computational and memory overhead. However, aggressive bitwidth reduction remains challenging due to activation outliers, leading to accuracy degradation. Existing methods, such as outlier-handling and group quantization, achieve high accuracy but incur substantial energy consumption. To address this, we propose SeVeDo, an energy-efficient SVD-based heterogeneous accelerator that structurally separates outlier-sensitive components into a high-precision low-rank path, while the remaining computations are executed in a low-bit residual datapath with group quantization. To further enhance efficiency, Hierarchical Group Quantization (HGQ) combines coarse-grained floating-point scaling with fine-grained shifting, effectively reducing dequantization cost. Also, SVD-guided mixed precision (SVD-MP) statically allocates higher bitwidths to precision-sensitive components identified through low-rank decomposition, thereby minimizing floating-point operation cost. Experimental results show that SeVeDo achieves a peak energy efficiency of 13.8TOPS/W, surpassing conventional designs, with 12.7TOPS/W on ViT-Base and 13.4TOPS/W on Llama2-7B benchmarks.

\end{abstract}

\begin{IEEEkeywords}
AI accelerators, heterogeneous architecture, group quantization, and mixed precision
\end{IEEEkeywords}

\section{Introduction}

\IEEEPARstart{R}{ecent} transformer models have achieved state-of-the-art results in language and vision tasks, but their growing size increases computation and memory overhead. Low-bit post-training quantization (PTQ) of weights and activations effectively mitigates these costs, enabling energy-efficient inference. However, transformers often exhibit extreme outliers in input activations, making them difficult to quantize effectively~\cite{awq,owq}. To address this, various outlier-handling and group quantization techniques have been proposed, exploiting the fact that outliers tend to appear in salient input channels~\cite{atom,edgediff,smoothquant,lightrot}. Yet these methods involve an inherent trade-off between accuracy and hardware cost, as finer control and granularity improve accuracy but incur substantial energy and area overhead~\cite{atom,tender,megamini}.

\begin{figure}[t]
    \centering
    \includegraphics[width=\linewidth]{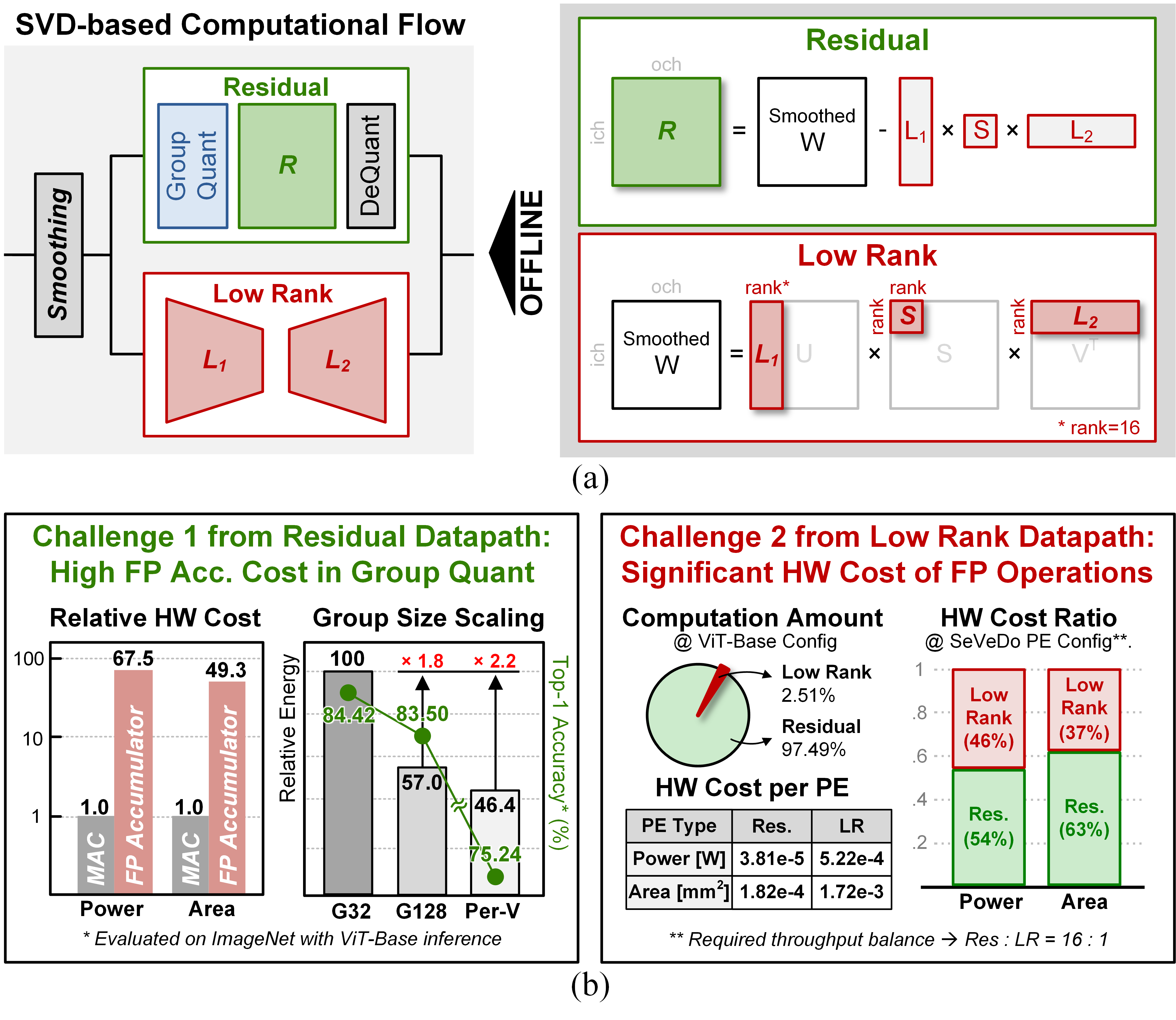}
    \caption{(a) SVD-based computational flow. (b) Challenges of SVD-based transformer architecture.}
    \label{fig_1}
    \vspace{-0.6cm}  
\end{figure}

The SVD-based computational flow, as illustrated in Fig.~1(a), achieves high accuracy with low-bit efficiency by applying SVD~\cite{svdquant} and group quantization. Truncated SVD is applied offline to decompose the weight matrix, structurally isolating salient components through the top-$k$ ranks (e.g., $k=16$) that capture the dominant portion of the weight with reduced dimensionality:
\[
W \approx U_k \Sigma_k V_k^T + R,
\]
The residual term $R$ retains the same dimension as the original weight but has its outliers removed by the low-rank decomposition, allowing it to be represented in a low-bit format. Also, group quantization~\cite{atom,edgediff,awq} can be orthogonally applied to further improve low-bit quantization accuracy by more precisely handling the remaining outliers.

Such hybrid processing introduces new hardware design challenges across each datapath, as depicted in Fig.~1(b). 1) Residual datapath: Finer group granularity increases the number of dequantization operations, where each floating-point (FP) accumulation involves costly multiplications between scaling factors and large-bitwidth partial sums—consuming up to 67.5$\times$ more power and 49.3$\times$ more area than an INT4 MAC. The overhead increases with smaller groups, diminishing hardware efficiency. 2) Low-rank datapath: Although the FP operations in the low-rank path~\cite{svdquant} account for only 2.51\% of total operations, it still occupies nearly half of the hardware cost—responsible for 46\% of power and 37\% of area under a PE configuration that preserves the minimum throughput balance. This hardware cost limits overall energy savings, as the FP path occupies a disproportionate share of total power.

Therefore, we propose SeVeDo, an energy-efficient SVD-based heterogeneous accelerator that addresses the aforementioned challenges through the following two key techniques. 

1) \textbf{Hierarchical Group Quantization (HGQ):} reduces FP accumulation cost via multi-scale quantization that combines coarse base and fine exponent-shifted scaling, achieving 36.1\% energy and 20.0\% area savings.  

2) \textbf{SVD-Guided Mixed Precision (SVD-MP):} applies mixed precision to statically identified precision-sensitive regions and executes them on bit-sliced INT units, yielding 75\% energy and 46\% area savings.

\section{Proposed Accelerator}

\subsection{Overall Architecture}

Fig. 2 illustrates the overall architecture of our accelerator optimized for statically-decomposed workloads. The system consists of four core clusters, each containing four heterogeneous cores sharing a 64KB IOMEM for activation management. All clusters are interconnected through a high-bandwidth NoC to a 1.5MB global memory, a top controller, and an auxiliary SIMD core. Each heterogeneous core integrates a Low-rank Vector Core (LVC) and a Residual Matrix Core (RMC) operating in parallel to balance high-precision, low-throughput computation with low-precision, high-throughput computation. The RMC employs an INT4 tensor PE array with a hierarchical quantization unit, whereas the LVC adopts a SIMD PE and bit-sliced datapath.

\begin{figure}[t]
    \centering
    \includegraphics[width=\linewidth]{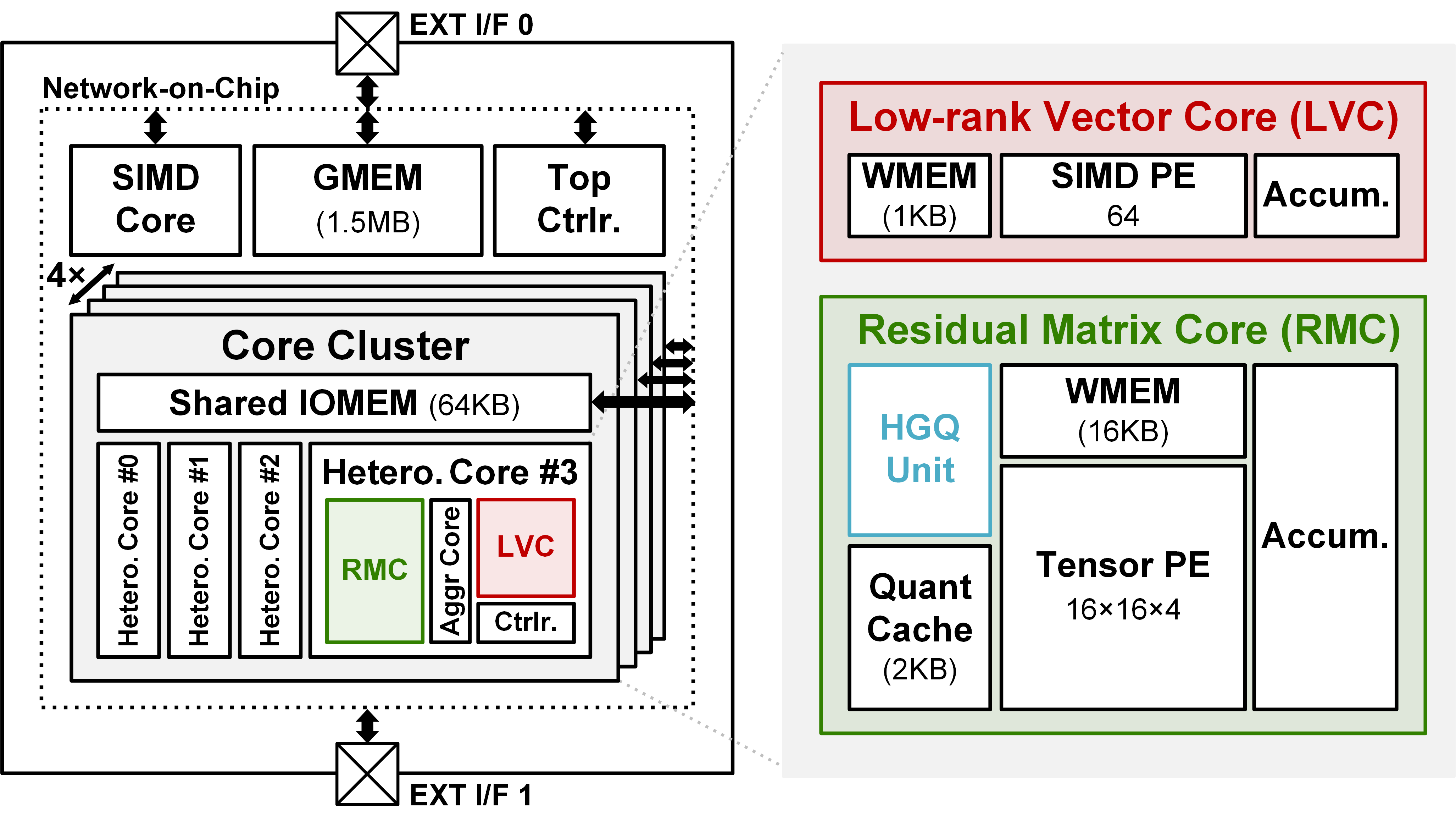}
    \caption{Overall architecture of SeVeDo.}
    \label{fig_2}
    \vspace{-0.6cm}
\end{figure}

\subsection{Hierarchical Group Quantization (HGQ)}

\begin{figure}[t]
    \centering
    \includegraphics[width=\linewidth]{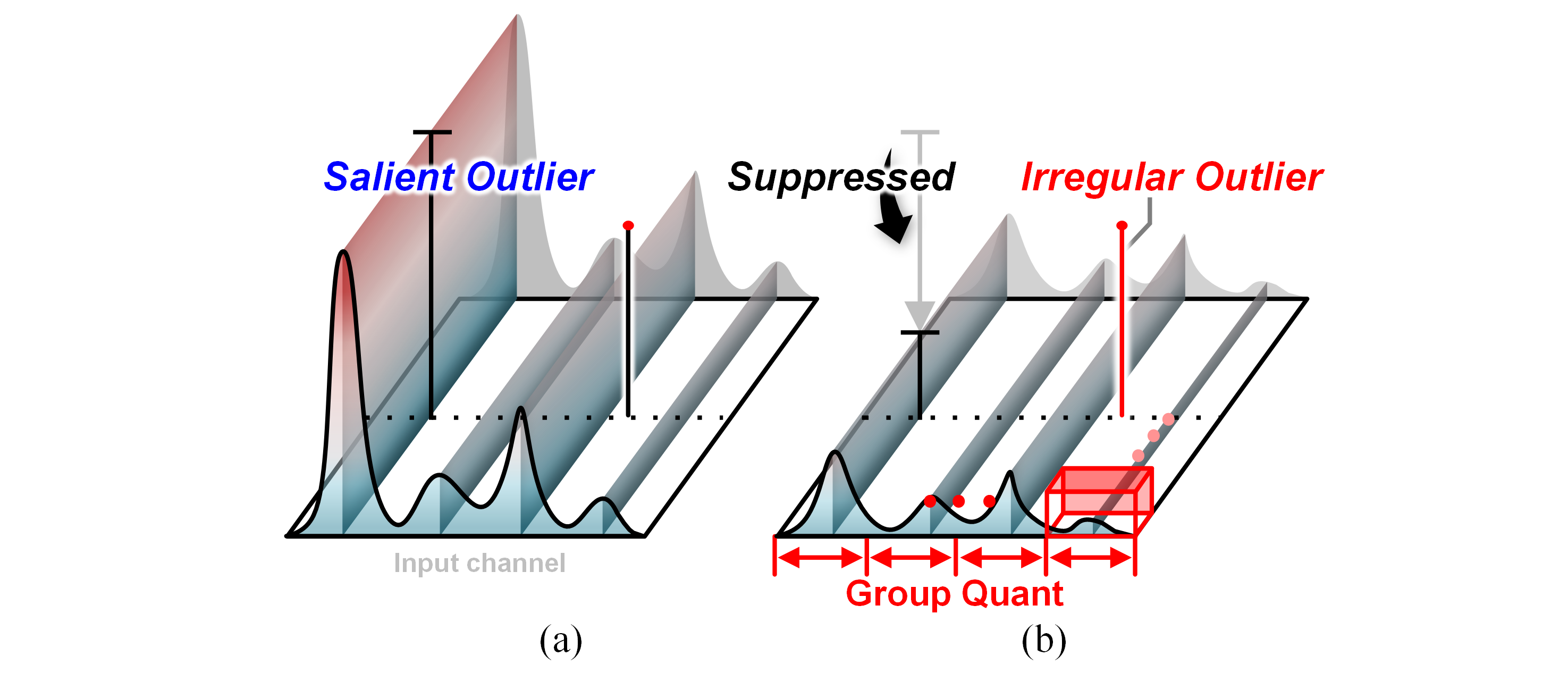}
    \caption{Motivation of HGQ. (a) Data distribution before SVD. (b) Residual processing with group quantization.}
    \label{fig_7}
    \vspace{-0.5cm}
\end{figure}

In the SVD-based dataflow, truncated SVD suppresses large activations in salient channels, enabling low-bit quantization of residual paths~\cite{smoothquant, svdquant}. However, since saliency is determined by the relative magnitude across channels, outliers in non-salient channels remain unsuppressed and become irregular, as illustrated in Fig.~3. These few residual outliers distort the global quantization range, degrading the effective resolution. Group quantization mitigates this by restricting the quantization range to local activation statistics, thereby improving suppression effectiveness. Despite its benefits, the group size introduces a critical trade-off between accuracy and hardware efficiency: coarse-grained quantization~\cite{atom} with accurate scaling factor per large group reduces hardware cost but loses precision, whereas fine-grained, exponent-only scaling schemes (e.g., MX~\cite{mx}, NVFP4~\cite{nvfp4}) sacrifice accuracy with slightly degraded resolution for cheaper integer-based multiplications.

\begin{figure}[t]
    \centering
    \includegraphics[width=\linewidth]{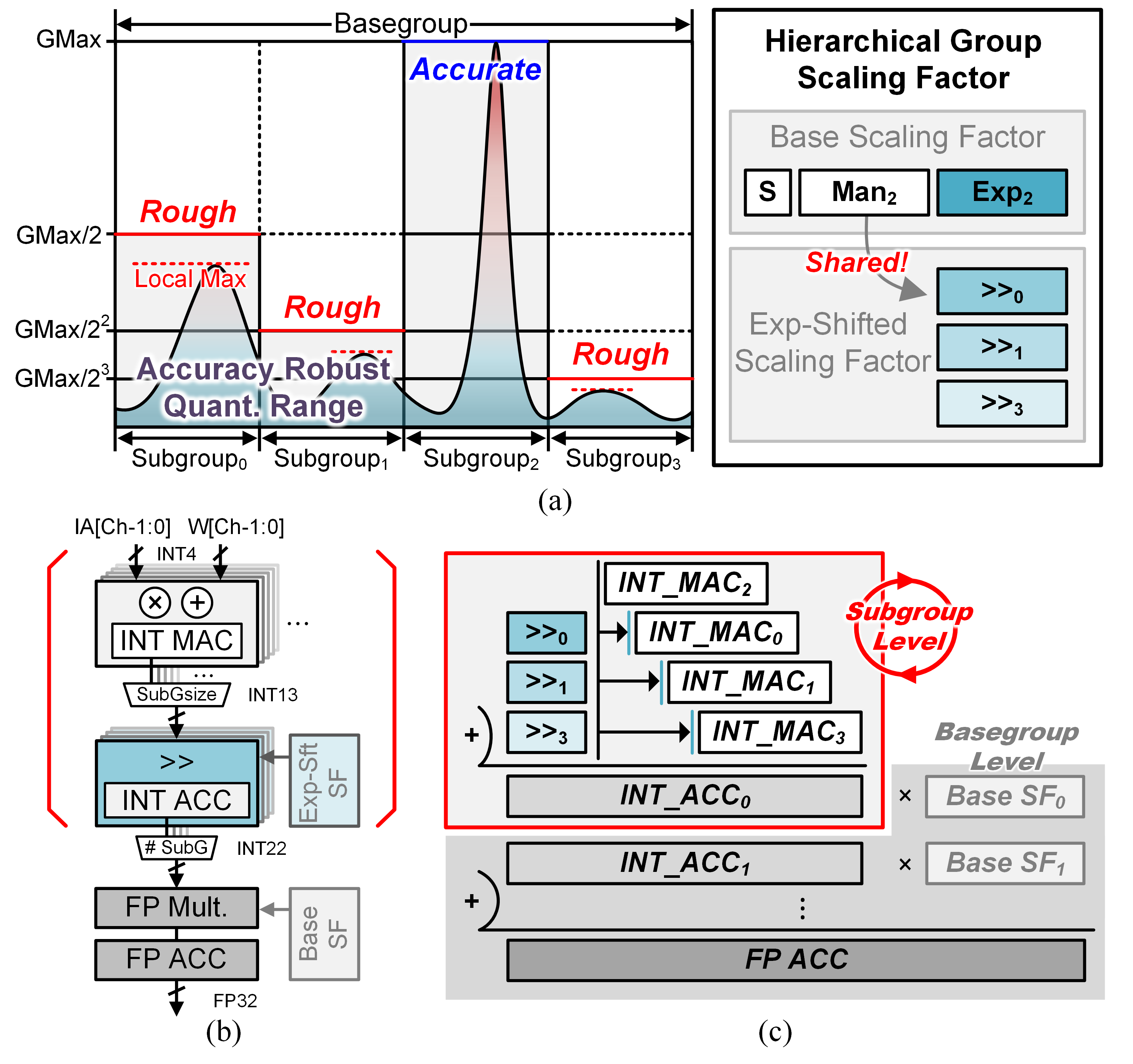}
    \caption{(a) Concept of HGQ. (b) Dequantization in tensor PE design. (c) High-level accumulation flow in subgroup level. }
    \label{fig_8}
    \vspace{-0.6cm}  
\end{figure}

To break this trade-off, we propose Hierarchical Group Quantization (HGQ), which employs a two-level scaling hierarchy—a base scaling factor (BSF) and an exponent-shifted scaling factor (ESSF)—to dynamically adjust quantization ranges. As shown in Fig. 4(a), after outlier suppression, sub-group maxima become much smaller than the base maximum, making them inherently tolerant to approximate scaling. This is because the upper bound of approximation error decreases logarithmically as the exponent shift increases. Each base group covers multiple sub-groups and is assigned a precise FP16 scaling factor derived from the global maximum, while each sub-group applies a lightweight exponent shift relative to BSF, aligning its local maximum with the nearest power-of-two level. This scheme reduces FP accumulation in proportion to the number of sub-groups, as shown in Fig.~4(b), while performing most accumulation in the integer domain through shift-based INT operations, as illustrated in Fig.~4(c). 

\begin{figure}[t]
    \centering
    \includegraphics[width=\linewidth]{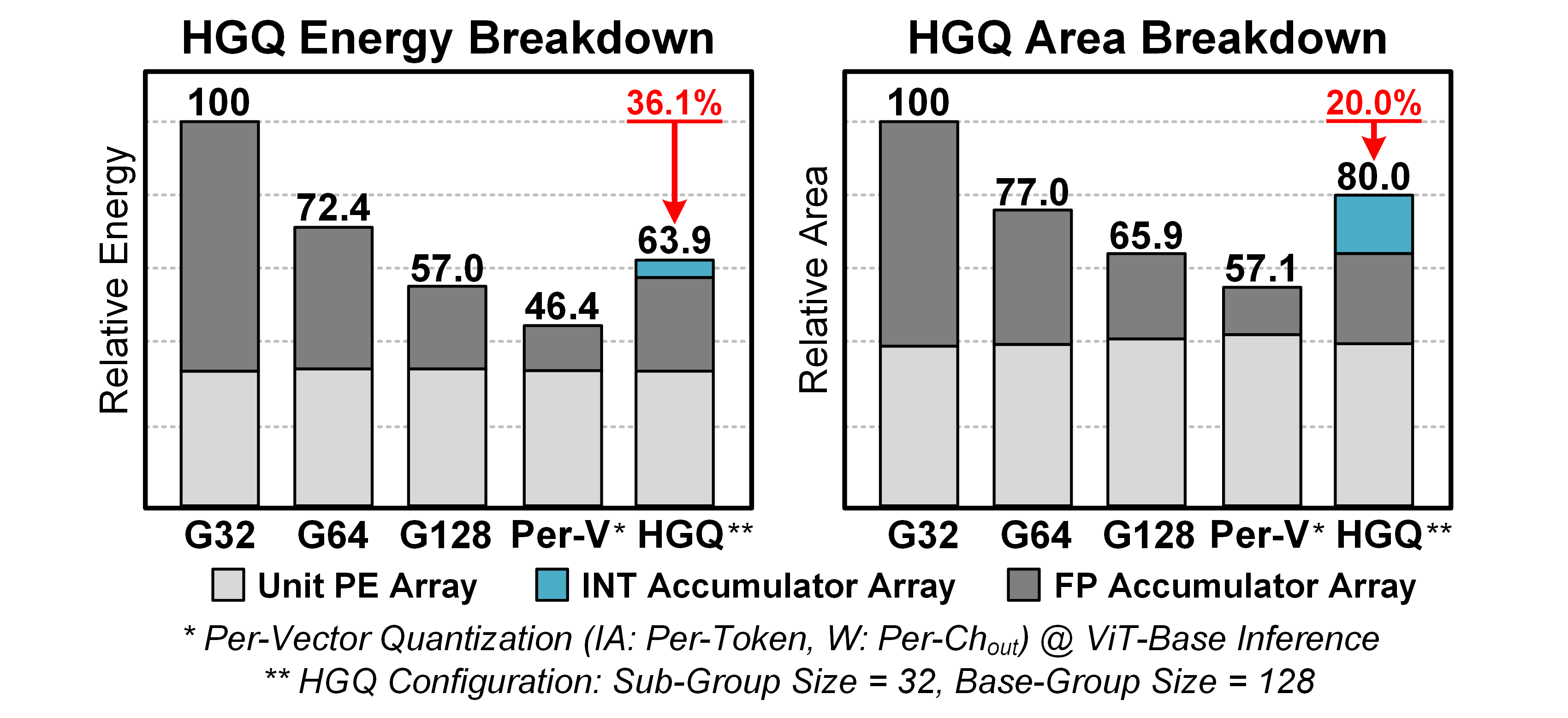}
    \caption{HGQ energy and area cost versus baseline quantization.}
    \label{fig_9}
    \vspace{-5pt}
\end{figure}

\begin{table}[t]
\centering
\caption{HGQ Accuracy on Transformer-Based Models}
\resizebox{\linewidth}{!}{
\begin{tabular}{cccccccccc}
\toprule
\multirow{3}{*}{\raisebox{0.45em}{\textbf{Method}}} 
& \multirow{3}{*}{\raisebox{1em}{\textbf{\makecell{Group\\Config.}}}}
& \multirow{3}{*}{\raisebox{1em}{\textbf{\makecell{Scaling\\Precision}}}}
& \multicolumn{3}{c}{\textbf{Llama$^{1)}$} (PPL↓)} 
& \multicolumn{3}{c}{\textbf{ViT$^{2)}$} (Accuracy↑)} \\
\cmidrule(lr){4-6} \cmidrule(lr){7-9}
& & & \makecell{1-7B\\[-2pt]} & \makecell{2-7B\\[-2pt]} & \makecell{2-13B\\[-2pt]} 
  & \makecell{Small\\[-2pt]} & \makecell{Base\\[-2pt]} & \makecell{Large\\[-2pt]} \\
\midrule
FP16        & --        & --       & 5.68      & 5.47      & 4.88      & 81.39     & 85.16     & 85.67    \\
\midrule
VS-Quant$^{3)}$\cite{vsquant} & G16/Per-V & INT4/FP16  & 10.47 & 9.56  & 20.44 & 79.49 & 83.52 & 85.34  \\
MXFP4\cite{mx}            & G32       & E8  & 7.23  & 7.53  & 6.22  & 52.20 & 64.39 & 79.92 \\
MXINT4\cite{mx}           & G32       & E8  & 7.07  & 7.52  & 5.87  & 75.38 & 82.33 & 84.87 \\
NVFP4\cite{nvfp4}         & G16/Per-T & FP8/FP32  & 6.27  & 6.14  & 5.34  & 77.62 & 83.33 & 85.17 \\
Tender$^{4)}$\cite{tender}   & Dyn/Per-T & E4/FP32  & 23.85 & 36.47 & 55.08 & --    & --    & --  \\
\midrule
\multirow{5}{*}{\raisebox{-0.9em}{INT4 w/o SVD}}
             & Per-V & FP16 & 4.4e3 & nan & 6.9e3 & 42.70   & 44.97   & 63.06    \\
             & G128  & FP16 & 6.93  & 6.79  & 5.81  & 72.04   & 81.80   & 84.75    \\
             & G64   & FP16 & 6.52  & 6.36  & 5.45 & 76.16   & 83.09   & 85.13    \\
             & G32   & FP16 & 6.29  & 6.13  & 5.28 & 78.04   & 83.78   & 85.36    \\
             & \makecell{HGQ\\(G32/G128)}  & E2/FP16 & 6.51  & 6.30  & 5.38 & 76.15 & 83.22 & 85.18 \\
\midrule
\multirow{5}{*}{\raisebox{-0.9em}{INT4 w/ SVD}}
             & Per-V & FP16 & 11.42 & 13.53 & 13.09 & 65.62 & 75.33 & 82.88    \\
             & G128  & FP16 & 6.25 & 6.09 & 5.33 & 76.71 & 83.43 & 85.35    \\
             & G64   & FP16 & 6.13 & 5.94 & 5.24 & 78.58 & 84.12 & 85.61    \\
             & G32   & FP16 & 6.03 & 5.84 & 5.15 & 79.37 & 84.27 & 85.64    \\
\rowcolor[HTML]{E0F2F5}
            & \textbf{\makecell{HGQ\\(G32/G128)}} & \textbf{E2/FP16} & \textbf{6.14} & \textbf{5.96} & \textbf{5.25} & \textbf{78.54} & \textbf{84.18} & \textbf{85.54}    \\
\bottomrule
\end{tabular}
}
\vspace{2pt}
\caption*{\scriptsize
1) Perplexity of the Llama-family models evaluated on Wikitext-2 \\
2) Top-1 accuracy of the ViT-family models evaluated on Imagenet-21k \\
3) Main precision INT4 with scaling factor Per-Group INT4/Per-Vector FP16 \\
4) Referenced from \cite{mant}; Main precision INT4; Not reproducible for ViT 
}
\vspace{-20pt}
\end{table}

Fig.~5 demonstrates the overall effectiveness of the proposed HGQ scheme. Our configuration adopts a sub-group size of G32 and a base-group size of G128, which means that 75\% of the FP accumulations in the G32 baseline are replaced with INT-domain accumulations. This substitution yields 36.1\% energy and 20.0\% area savings compared to the G32 configuration.

Table~I further details the accuracy comparison with prior multi-scale quantization methods~\cite{vsquant, tender, nvfp4, mx} under various design points. Regardless of whether SVD is applied, HGQ consistently achieves higher accuracy compared to the G128 baseline while maintaining a comparable hardware cost. In addition, unlike~\cite{vsquant, mx}, which use small group sizes (G16–G32) with low scaling precision (INT4/E8), HGQ effectively maintains high scaling precision through its hierarchical scaling structure, outperforming them. Furthermore, while NVFP4~\cite{nvfp4} employs FP8 scaling factors with G16 groups at significantly higher cost, our design combines SVD and HGQ to achieve even higher accuracy at lower cost.

\subsection{SVD-Guided Mixed Precision (SVD-MP)}

\begin{figure}[t]
    \centering
    \includegraphics[width=\linewidth]{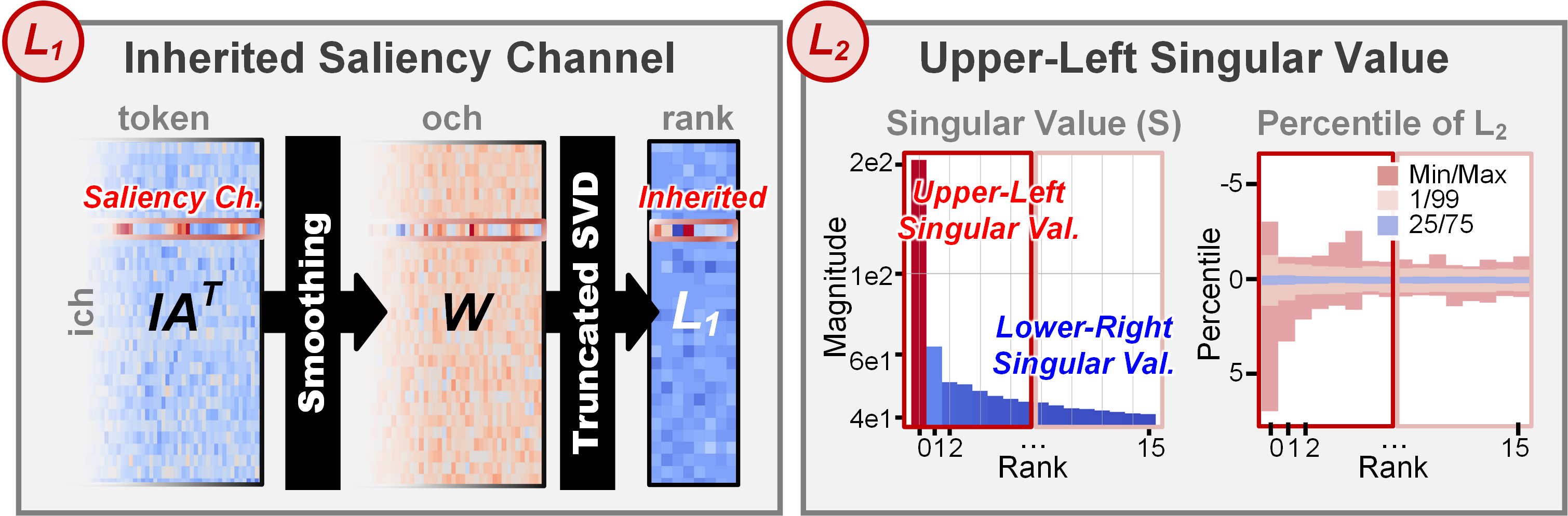}
    \caption{Precision sensitive regions after SVD: Inherited outlier in L$_1$ projection, upper-left singular values in L$_2$ projection.}
    \label{fig_11}
    \vspace{-0.2cm}  
\end{figure}

The baseline implementation~\cite{svdquant} adopts high-precision FP16 computations for both the L$_1$ and L$_2$ projection matrices in the low-rank path. However, as discussed in Section I, this path still consumes nearly half of the overall power and area despite its limited operation count. As depicted in Fig.~6, leveraging truncated SVD enables static handling of outliers, allowing us to identify precision-sensitive regions while providing opportunities to compute less critical regions at lower cost. The first precision-sensitive region originates from inherited saliency channels in the L$_1$ projection layer; these channels often maintain large magnitudes even after low-rank truncation. The second region appears in the upper-left corner of the singular value matrix, where the singular values are sorted in descending order. This region corresponds to the upper input channels of the activation and L$_2$ weight matrices, amplifying their magnitudes.

\begin{figure}[t]
    \centering
    \includegraphics[width=\linewidth]{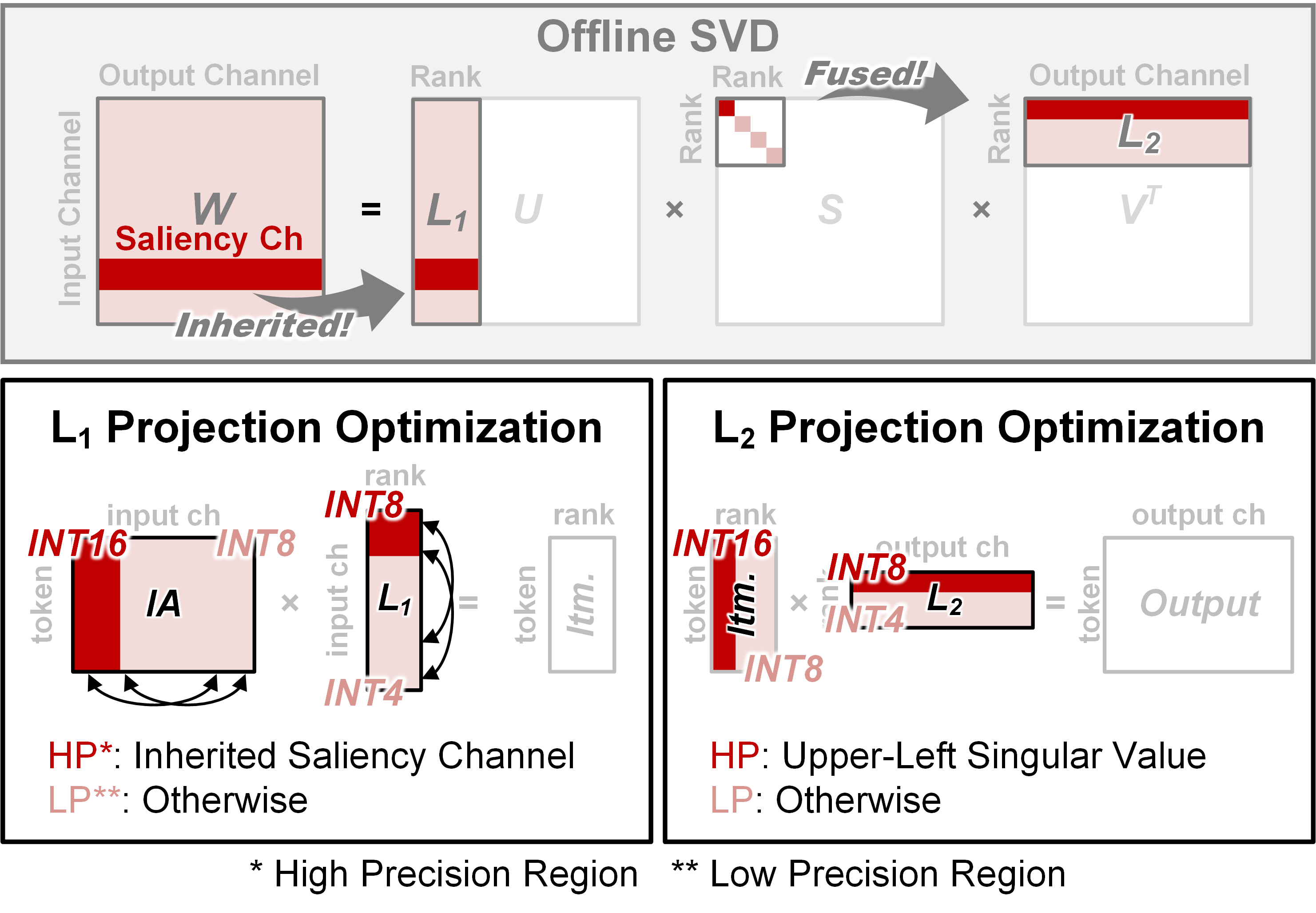}
    \caption{SVD-MP scheme: inherited saliency channels and singular-value-aware projection optimizations for L$_1$ and L$_2$.}
    \label{fig_12}
    \vspace{-0.5cm}  
\end{figure}

Fig.~7 shows the details of the SVD-MP operation. First, the top-128 and top-4 precision-sensitive channels are identified in the L$_1$ and L$_2$ projections, respectively. Then, mixed-precision is applied by assigning higher precision only to these regions in both activations and weights. Weights are reordered and quantized offline, using INT8 for sensitive channels and INT4 for the others. Activations are processed online by exponent-aligning inputs and assigning INT16 to sensitive channels and INT8 to the rest. To realize this algorithm, we design a hardware architecture that supports time-multiplexed bit-slice processing. In this structure, INT16–INT8 and INT8–INT4 operations are executed sequentially within the same compute flow, requiring a temporally reconfigurable bit-slice PE. As shown in Fig. 8, the PE dynamically switches its operation mode along the time axis, allowing the same unit to alternately process multiple precisions along with the corresponding exponent maxima and shift amounts. During the high-precision phase, the PE accumulates four MSB–LSB slice combinations, while in the low-precision phase, it completes the computation in a single cycle. All operands, including MSB/LSB slices and low-precision inputs, are sign-extended for consistent signed arithmetic.

\begin{figure}[t]
    \centering
    \includegraphics[width=\linewidth]{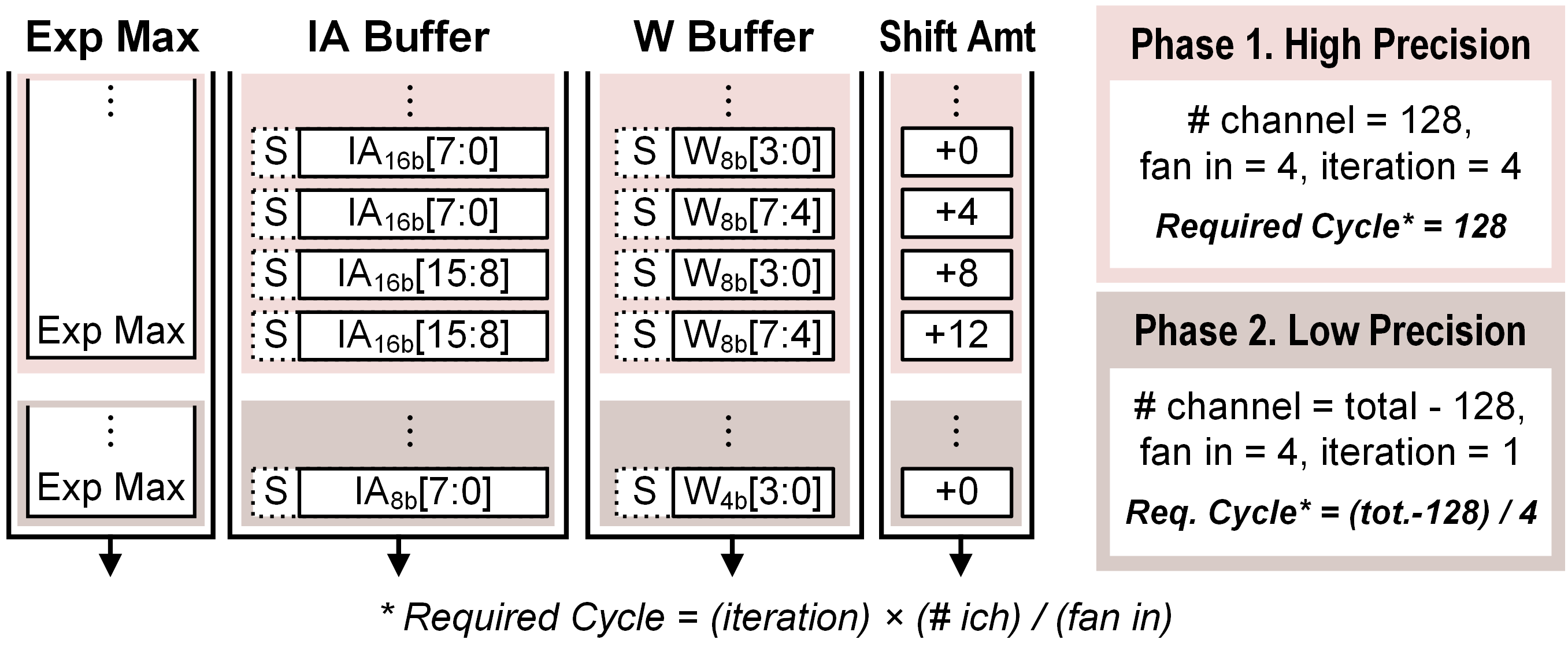}
    \caption{SVD-MP data preprocessing and feeding strategy.}
    \label{fig_13}
    \vspace{-0.2cm}  
\end{figure}

\begin{figure}[t]
    \centering
    \includegraphics[width=\linewidth]{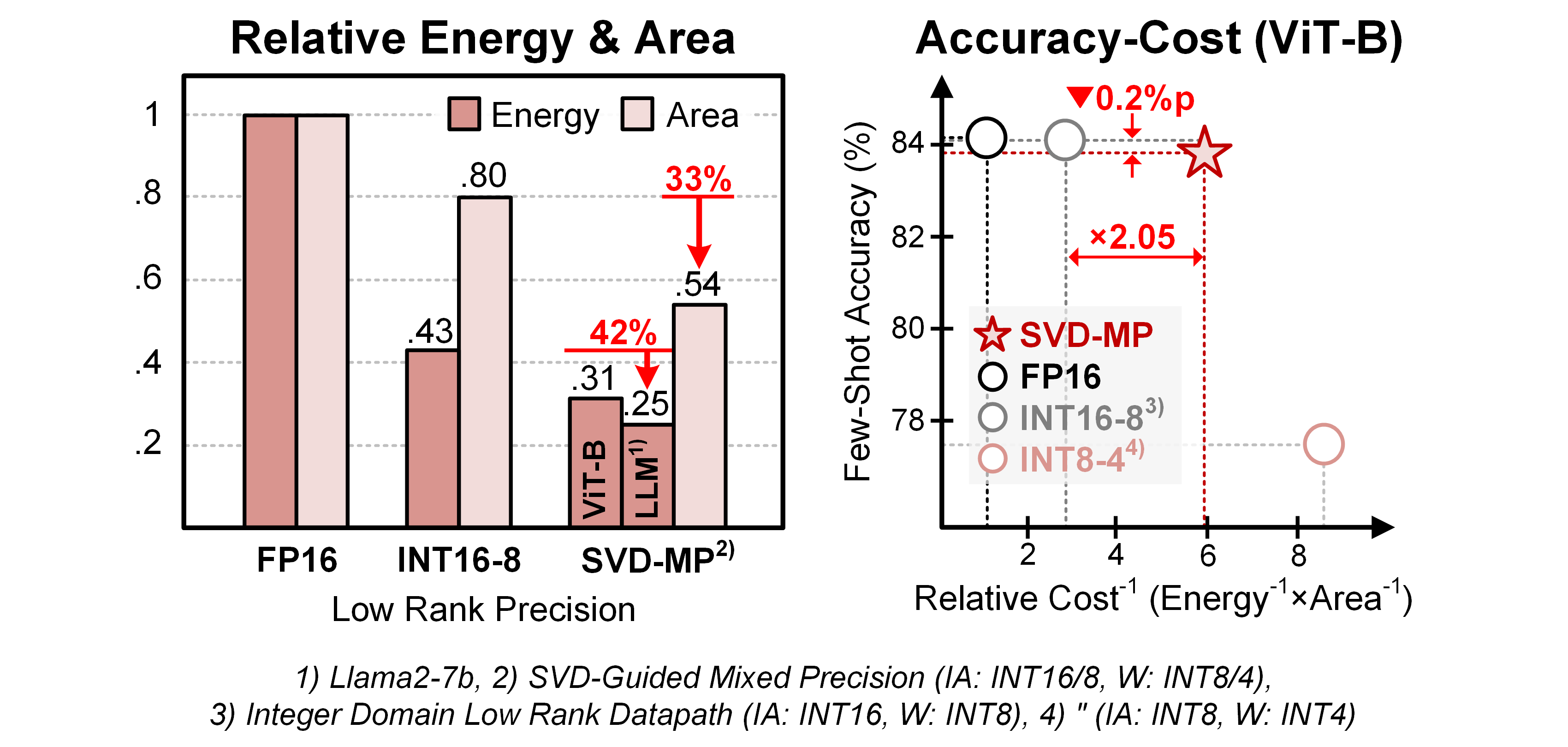}
    \caption{Accuracy–cost trade-off of SVD-MP.}
    \label{fig_14}
    \vspace{-0.5cm}  
\end{figure}

As summarized in Fig.~9, SVD-MP reduces both energy and area in the low-rank datapath compared to floating-point and high-bit integer baselines. The left figure presents the relative energy and area results, showing that our mixed-precision scheme achieves 33\% area and up to 42\% energy reduction over the INT16–8 baseline for Llama2-7B and ViT-Base inference. This corresponds to only 25\% of the energy and 54\% of the area compared to the FP16 baseline adopted in the original implementation~\cite{svdquant}. The right figure illustrates the accuracy–cost comparison for ViT-Base. SVD-MP achieves 2.05× higher hardware efficiency than the INT16–INT8 baseline while maintaining accuracy loss within 0.2\%p. In contrast, the INT8–INT4 configuration exhibits larger cost reduction but suffers from a notable accuracy drop, highlighting the superior efficiency–accuracy balance achieved by SVD-MP.

\section{Implementation Results}

\begin{figure}[t]
    \centering
    \includegraphics[width=\linewidth]{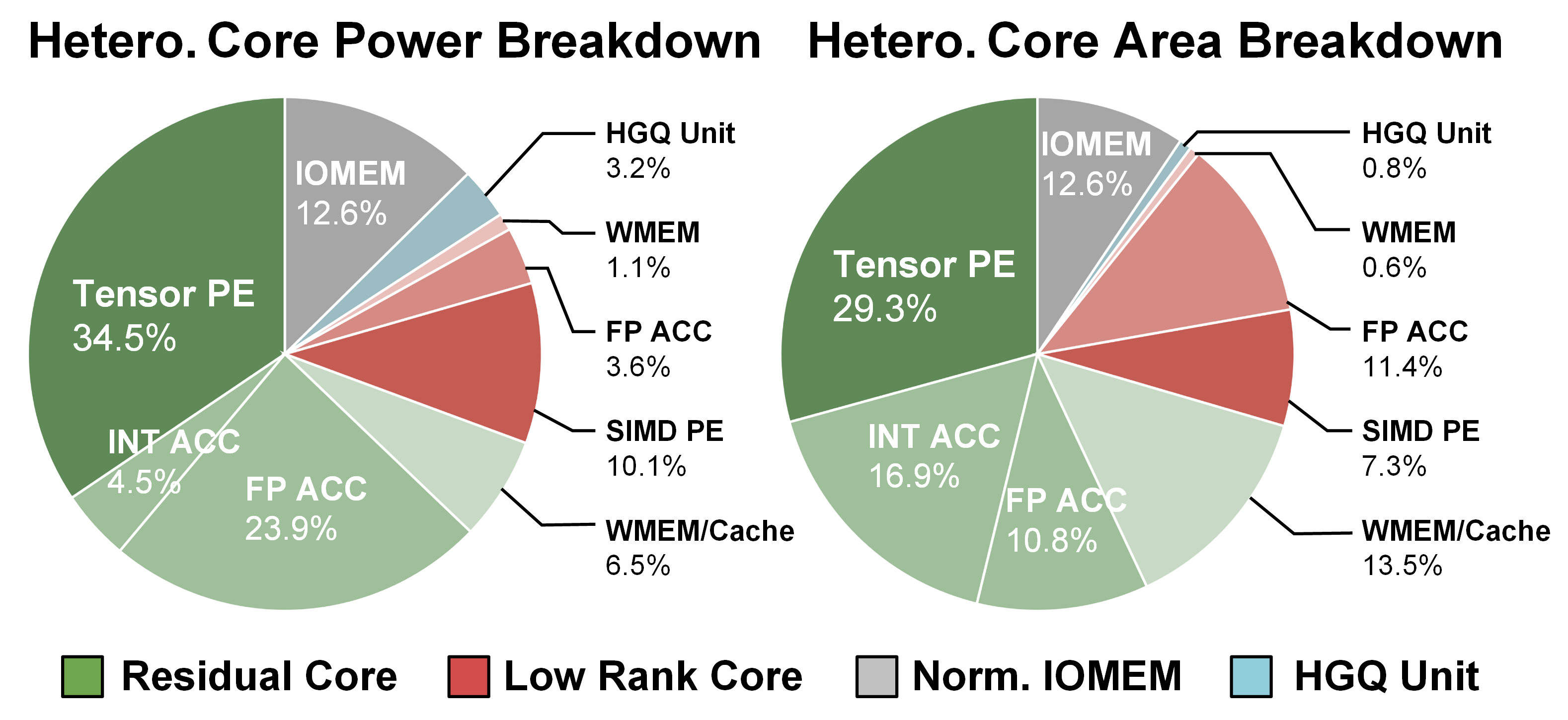}
    \caption{Power and area breakdown of Heterogeneous Core.}
    \label{fig_15}
    \vspace{-0.2cm}  
\end{figure}

Our accelerator is implemented in Samsung 28nm CMOS process with a die area of 10.72mm². It operates at 250MHz under a 0.9V supply, achieving an energy reduction of 54\% compared to the baseline design. As summarized in Table II, the proposed heterogeneous architecture effectively handles outliers through SVD-based decomposition and fine-grained group quantization, leveraging both static and dynamic saliency channel characteristics from language and vision tasks. Compared with previous works~\cite{tambe, ayaka, megamini}, our design achieves higher energy efficiency with accurate low-bit operation enabled by SVD-based decomposition. Although~\cite{megamini} also demonstrates high peak efficiency, its benchmark efficiency degrades significantly in practice due to the high proportion of outliers in real workloads. In contrast, the proposed SeVeDo architecture performs optimized outlier handling through SVD-based decomposition and maintains high overall efficiency through heterogeneous optimization across the residual and low-rank paths.

\begin{table}[t]
    \centering
    \caption{Comparison Table}
    \includegraphics[width=\linewidth]{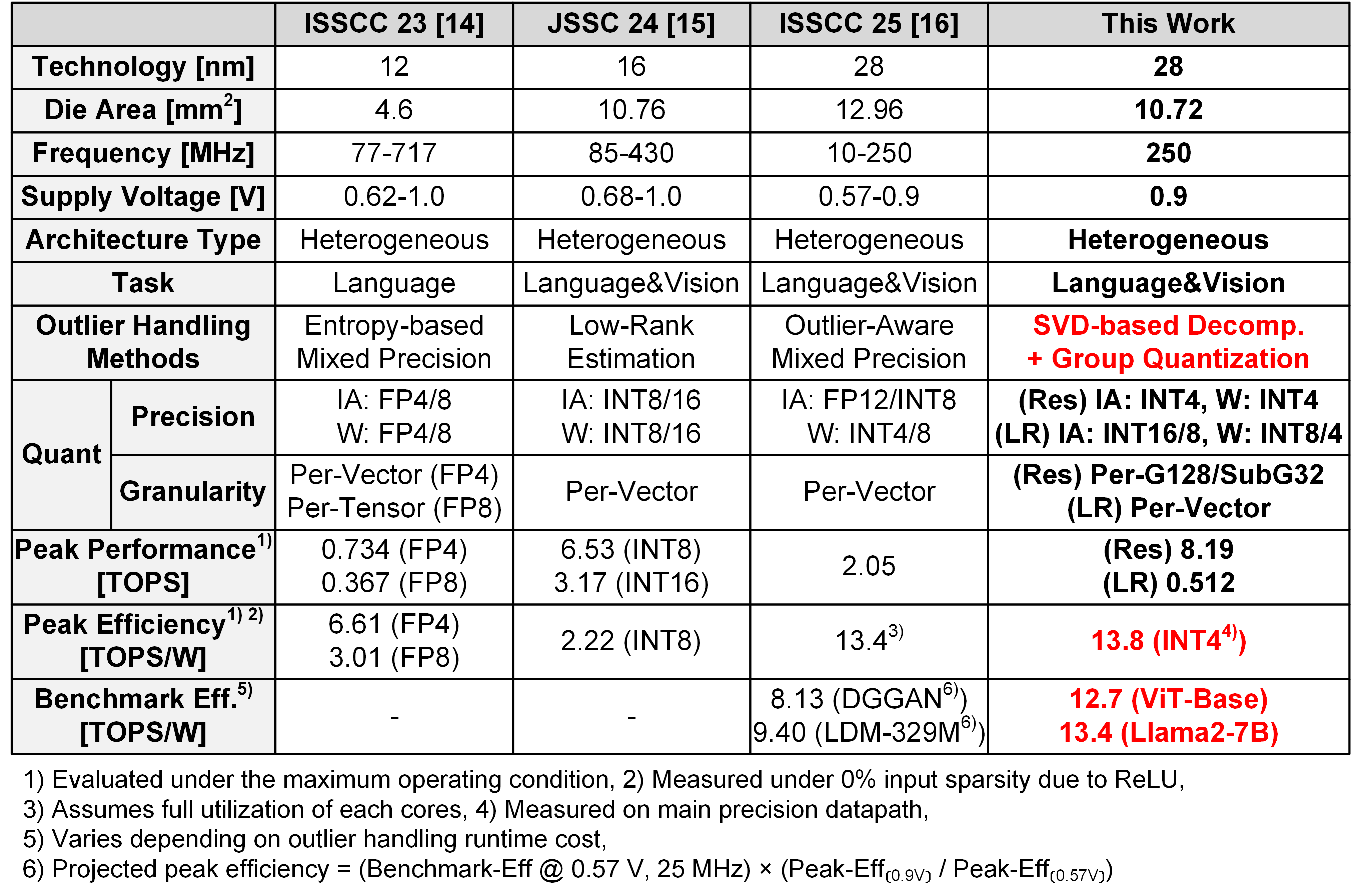}
    \label{tab:1}
    \vspace{-1.0cm}
\end{table}

\section{Conclusion}

The proposed SeVeDo accelerator achieves energy-efficient transformer inference by leveraging the distinct characteristics of residual and low-rank datapaths introduced by SVD decomposition. The heterogeneous architecture incorporates two core-level innovations to efficiently support these decomposed datapaths: (1) Hierarchical Group Quantization (HGQ) reduces FP accumulation overhead in the residual path by replacing 75\% of FP accumulations with shift-based INT operations, achieving 36.1\% energy and 20.0\% area reduction over the G32 baseline; and (2) SVD-Guided Mixed Precision (SVD-MP) exploits the structural properties of SVD for integer-domain mixed precision, reducing high-cost FP operations and achieving 75\% energy and 46\% area savings compared to the FP16 baseline. As a result, SeVeDo delivers a benchmark energy efficiency of 12.7–13.4TOPS/W on ViT-Base and Llama2-7B, while demonstrating a superior efficiency–accuracy trade-off over previous implementations.

\vspace{12pt}

\end{document}